\documentclass[12pt]{elsarticle}
\usepackage{amssymb}
\usepackage{times}
\usepackage{latexsym}
\usepackage{graphicx}
\usepackage{caption}
\usepackage{subcaption}
\usepackage[T1]{fontenc}
\usepackage[utf8]{inputenc}
\usepackage{xcolor}
\usepackage{url}

\journal{Natural Language Processing Journal}

\begin{document}
\setcitestyle{square}
\begin{frontmatter}


 
\title{Incorporating Dictionaries into a Neural Network Architecture to Extract COVID-19 Medical Concepts From  Social Media}
\author[mymainaddress]{Abul Hasan \corref{mycorrespondingauthor}}
\cortext[mycorrespondingauthor]{Corresponding author}\ead{a.hasan@bbk.ac.uk}
\author[mymainaddress,mythirdaddress]{Mark Levene}
\author[mymainaddress]{David Weston}
\address[mymainaddress]{Birkbeck, University of London, Department of Computer Science and Information Systems, London WC1E 7HX, UK}
\address[mythirdaddress]{Department of Data Science, National Physical Laboratory (NPL) Hampton Road, Teddington, London TW11 0LW, UK}

\begin{abstract}
We investigate the potential benefit of incorporating dictionary information into a neural network architecture for natural language processing. In particular, we make use of this architecture to extract several concepts related to COVID-19 from an on-line medical forum. We use a sample from the forum to manually curate one dictionary for each concept. In addition, we use MetaMap, which is a tool for extracting biomedical concepts, to identify a small number of semantic concepts. For a supervised concept extraction task on the forum data, our best model achieved a macro $F_1$ score of 90\%. A major difficulty in medical concept extraction is obtaining labelled data from which to build supervised models. We investigate the utility of our models to transfer to data derived from a different source in two ways. First for producing labels via weak learning and second to perform concept extraction.  The dataset we use in this case comprises COVID-19 related tweets and we achieve an $F_1$ score 81\% for symptom concept extraction trained on weakly labelled data. The utility of our dictionaries is compared with a COVID-19 symptom dictionary that was constructed directly from Twitter. Further experiments that incorporate BERT and a COVID-19 version of BERTweet demonstrate that the dictionaries provide a commensurate result. Our results show that incorporating small domain dictionaries to deep learning models can improve concept extraction tasks. Moreover, models built using dictionaries generalize well and are transferable to different datasets on a similar task.
\end{abstract}



\end{frontmatter}


\section{Introduction}
\label{sec:intro}
Extracting COVID-19 symptoms both from social media and from medical documents has been found to be useful for tracking this disease \cite{klein2021toward} and for building prognosis models to predict mortality in hospitals \cite{silverman2021nlp}.
\par There are many approaches to medical concept extraction, for example \cite{hasan2022monitoring} successfully used a CRF that included manually created dictionary features to extract medical concepts from social media. In contrast a Long Short-Term Memory (LSTM) network initialised with word and character embeddings \cite{lample-etal-2016-neural} demonstrated improvements in concept extraction over a feature based CRF from a variety of formal document datasets \cite{habibi2017deep}. More recently, Bidirectional Encoder Representations from Transformers (BERT) \cite{devlin-etal-2019-bert} has achieved state of the art performances in a variety of NLP tasks including biomedical text mining \cite{lee2020biobert}. However, one of the bottlenecks for deploying such architectures is the need for labelled data. Recent work in medical concept extraction such as in \cite{sun-bhatia-2021-neural}, has utilised dictionaries/gazetteers with deep learning models to leverage external knowledge when labelled datasets are scarce.
\begin{figure*}[!h]
	\centering
	\begin{subfigure}{.5\textwidth}
		\centering
		\includegraphics[scale=0.9]{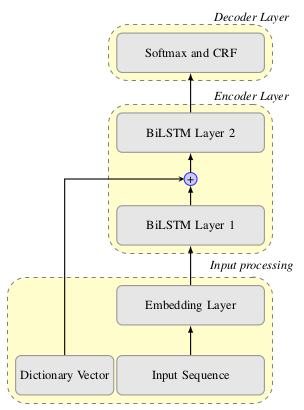}
		\caption{LSTM+CRF architecture}
		\label{fig:01_a}
	\end{subfigure}%
	\begin{subfigure}{.5\textwidth}
		\centering
		\includegraphics[scale=0.9]{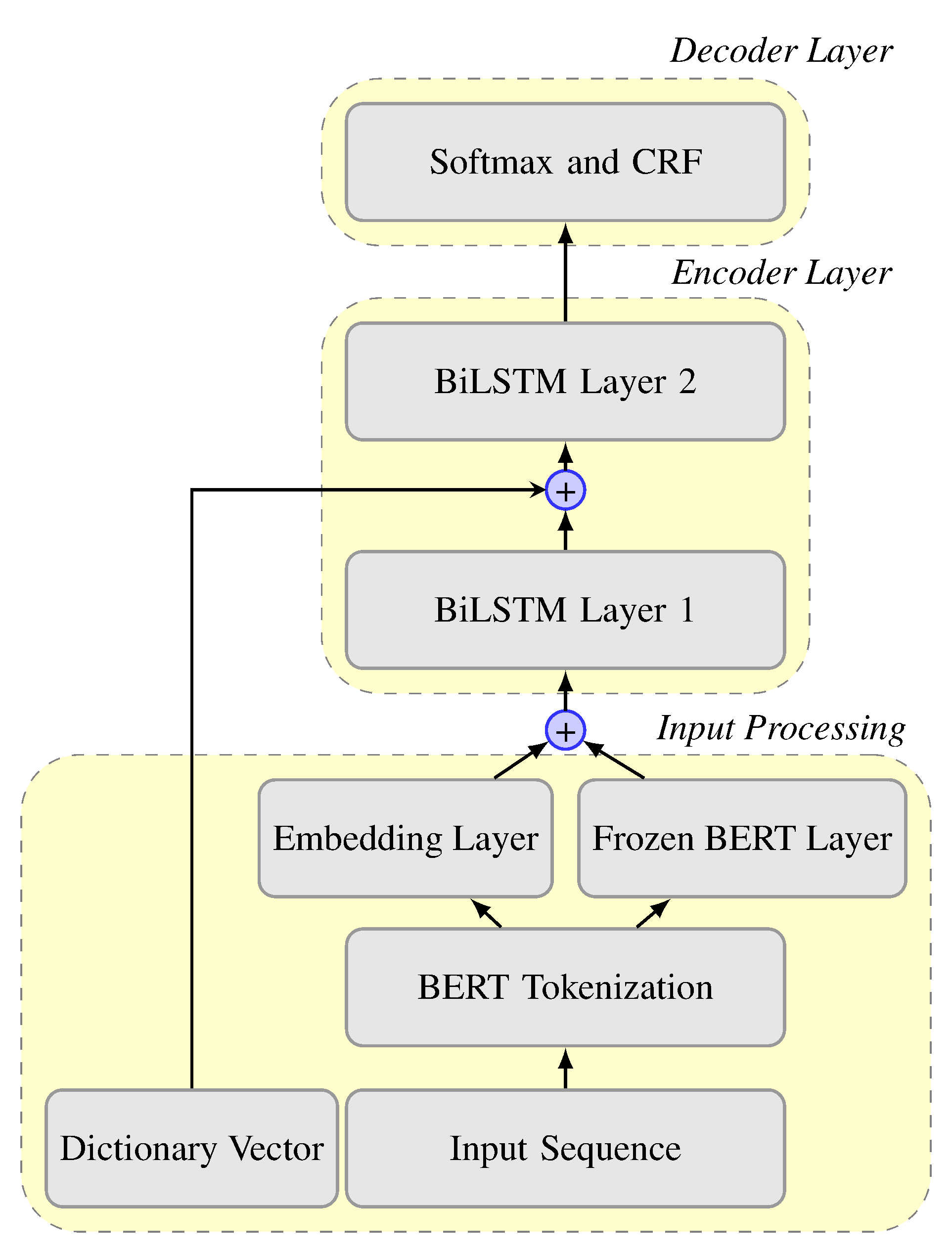}
		\caption{BERT+LSTM+CRF architecture}
		\label{fig:01_b}
	\end{subfigure}
	\caption{ Neural architectures for extracting COVID-19 medical concepts from social media. In both architectures shown the dictionary information is appended to the input of layer 2.}
	\label{fig:01}
\end{figure*}
\par Herein, we investigate two variants of an LSTM based deep learning architecture that essentially differ in the type of input provided. In Figure ~\ref{fig:01}a. the input sequence is replaced with vectors from a pre-trained static word embedding. In Figure ~\ref{fig:01}b the input static word embedding vector  is concatenated with a vector from a contextual word embedding. We are interested in the effect of including dictionary features to these models. In this work we investigate the effect of appending a dictionary vector (which we describe in Section 3) to either the input of the first or second BiLSTM layer. The figures show the case where dictionary vector is injected into the input of Layer 2. The investigation is performed using a COVID-19 discussion forum \cite{PI} dataset which was annotated with several medical concepts (see Dataset description). The results show that models built incorporating dictionaries perform better than the those without them. Furthermore, in order to check the transferability of the models and dictionaries to a publicly available COVID-19 Twitter dataset \cite{chen2020tracking}, weak supervision methodology is developed. Specifically, we utilise the manually built symptom dictionary from \cite{hasan2022monitoring} and a publicly available COVID-19 symptom dictionary from \cite{sarker2020self}. These are termed as (i) {\em Our dictionary}, (ii) {\em Sarker dictionary}, and (iii) {\em Combined dictionary}. The combined dictionary is the merger of former two dictionaries. First, two base-line models are trained using labelled dataset by Our and Sarker dictionaries separately. Then the models are retrained by incrementally adding terms from either dictionaries; i.e. the coverage of Our dictionary is increased by Sarker's and vice-versa. These models are tested with the dataset tagged by Our, Sarker, and Combined dictionaries, respectively. Furthermore, they are tested with a manually annotated ground truth data. For the Twitter dataset, we use COVID-19 version of BERTweet \cite{nguyen-etal-2020-bertweet}, and for the forum dataset we use BERT base model \cite{devlin-etal-2019-bert}.\\
Our contributions are as follows:
\begin{enumerate}
	\item With combination of static and contextual word embeddings and by leveraging dictionary features, we obtain a very good performance in extracting COVID-19 medical concepts from social media text. 
	\item We show that dictionaries are useful as weak learners and the neural model achieve a very good performance when we transfer it to extract COVID-19 symptoms from a larger Twitter dataset.  
\end{enumerate}
\begin{figure*}[!ht]
	\centering
	\includegraphics[scale=0.9]{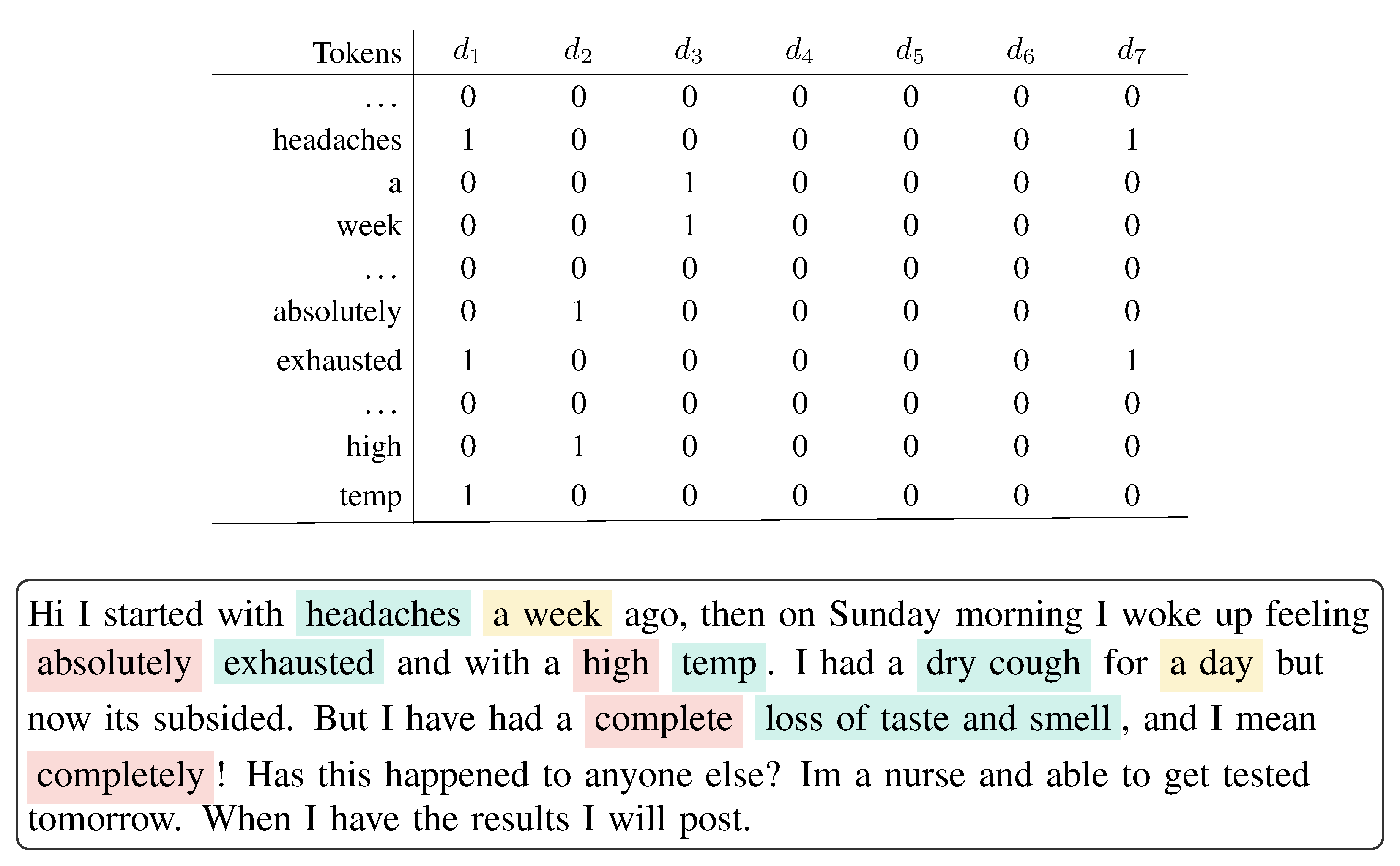}
	\caption{An example post and its feature matrix for a selected sequence. Here, green, yellow, and red denote symptom, duration, and severity concepts. Moreover, $d_1$ to $d_6$ denote symptom, severity, duration, intensifier, negation, and body parts dictionaries, respectively, and $d_7$ represents MetaMap. }
	\label{fig:02}
\end{figure*}
\section{Related Works}
\label{sec:rel}

Social media is a valuable source for monitoring public health information such as detecting influenza epidemics \cite{aramaki2011twitter}, and finding adverse drug reactions (ADR) \cite{nikfarjam}. The early work of Karimi et al. \cite{karimi2015cadec} and Jimeno-Yepes \cite{jimeno2015identifying} are notable since they published two social media corpora named CSIRO Adverse Drug Event Corpus(CADEC) and Micromed, respectively. Both corpora are annotated with medical concepts such as ADR, symptoms, and drugs form two sources;  online-forum posts and  Twitter. However, their coverage was  limited to 1321 posts and 1300 tweets only. Since its inception in 2016, the Social Media Mining for Health Research and Applications (SMM4H) shared tasks have been a regular venue for publishing larger annotated datasets related to diverse categories of tasks. These tasks include but not limited to ADR detection \cite{sarker2016social, sarker2017overview, weissenbacher-etal-2019-overview, klein-etal-2020-overview}, automatic classification of tweets mentioning a drug name \cite{weissenbacher-etal-2018-overview}, and  more recently, the classification of tweets containing COVID-19 symptoms \cite{magge-etal-2021-overview}. \\
Though extraction of medical concepts from social media is a challenging task due to the possible use of colloquial language, with the advent of deep learning such concept extraction tasks have seen a  significant performance improvement. For example, Cocos et al. \cite{cocos2017deep} first applied a BiLSTM architecture for extracting ADR from the dataset published by \cite{nikfarjam} and reported to achieve a $F_1$ score of 75.5\%. Scepanovic et al. \cite{scepanovic2020extracting} extracted medical concepts from CADEC and Micromed and achieved 82\% and 72\%, respectively, by deploying a BiLSTM-CRF architecture and combining different word embeddings and BERT. An attention mechanism was  also used with a BiLSTM-CRF architecture for the same task in \cite{chowdhury2018multi, zhang2019adverse}. In the case of the SMM4H shared task for ADR extraction, transformer based models have recently dominated \cite{portelli-etal-2021-bert}. Specifically, Dima et al. \cite{dima-etal-2021-transformer} framed the ADR extraction as a multi task detection task using BioBERT \cite{lee2020biobert} and achieved the best performance for ADR extraction task to  date.
Since the start of the COVID-19 pandemic, social media data such as Twitter have been extensively used to track and monitor COVID-19 outbreaks and novel symptoms \cite{klein2021toward, guo2020mining, sarker2020self}. Specifically, Sarker et al. \cite{sarker2020self} published a COVID-19 symptom dictionary comprising colloquial terms used in a Twitter dataset. Hernandez et al. \cite{hernandez2021biomedically} explored a large scale Twitter dataset  \cite{banda2021large} and automatically tagged tweets for drugs, conditions/symptoms, and measurements using biomedical taggers such as ScispaCy \cite{neumann-etal-2019-scispacy}. They found that existing biomedical/scientific text processing systems for concept extraction do not generalize well when used with non-clinical data sources like Twitter \cite{hernandez2021biomedically}.
\par
Recent works on incorporating gazetteers/lexicons into neural models focus on creating gazetteer embeddings and gazetteer models \cite{song2020improving,  peshterliev2020self, magnolini-etal-2019-use}. For example, Sun et al. \cite{sun-bhatia-2021-neural} created gazetteer embeddings using labels and related gazetteers and fused them to a BERT-based encoder. 
\par
Weak supervision is a process by which labelling functions are used to automatically annotate unlabelled data. For example, in the biomedical domain, SwellShark \cite{fries2017swellshark} uses lexicons from different sources and rules to generate labelled dataset for training with BiLSTM-CRF network. Recently, the WRENCH \cite{zhang2021wrench} benchmark has come up with a set of generalised functions to programmatically produce labels for a diverse set of datasets. 
\begin{table*}[htb!]
	
	\begin{center}
		\begin{tabular}{ c|ccc|ccc|ccc}
			&\multicolumn{3}{c}{\textbf{LSTM+CRF}}&\multicolumn{3}{c}{\textbf{+DICT(1)}}&\multicolumn{3}{c}{\textbf{+DICT(2)}} \\ \cline{2-10}   
			Label&P&R&$F_1$&P&R&$F_1$&P&R&$F_1$\\ \hline
			SYM&0.84&0.77&0.80&0.92&0.94&0.93&0.93&0.95&0.94\\ 
			SEVERITY&0.67&0.51&0.58&0.74&0.77&0.75&0.75&0.80&0.77\\ 
			BPOC&0.82&0.89&0.85&0.91&0.88&0.89&0.90&0.90&0.90\\ 
			INTENSIFIER&0.82&0.90&0.86&0.87&0.94&0.91&0.88&0.94&0.91\\ 
			DURATION&0.79&0.79&0.79&0.81&0.91&0.86&0.85&0.89&0.87\\ 
			NEGATION&0.81&0.88&0.84&0.83&0.83&0.83&0.81&0.87&0.84\\
			O&0.96&0.97&0.97&0.98&0.98&0.98&0.99&0.98&0.98\\ \hline
			MACRO&0.82&0.82&0.81&0.87&0.89&0.88&0.87&0.90&\textbf{0.89}\\ \hline 	
		\end{tabular}
	\end{center}
	\caption{Results of concept extraction from forum dataset using {\textbf{LSTM+CRF}} architecture. For the descriptions of {\textbf{LSTM+CRF}}, {\textbf{+DICT(1)}}, and {\textbf{+DICT(2)}} models see Subsection ~\ref{subsec:archmodels}} \label{tab:01}
\end{table*}
\begin{table*}[h!]
	
	\begin{center}
		\begin{tabular}{ c|ccc|ccc|ccc}
			&\multicolumn{3}{c}{\textbf{BERT+LSTM+CRF}}& \multicolumn{3}{c}{\textbf{+DICT(1)}}&\multicolumn{3}{c}{\textbf{+DICT(2)}}\\ \cline{2-10}  
			Label&P&R&$F_1$&P&R&$F_1$&P&R&$F_1$\\ \hline
			SYM&0.79&0.86&0.82&0.92&0.92&0.92&0.93&0.97&0.95\\ 
			SEVERITY&0.70&0.38&0.49&0.75&0.66&0.69&0.75&0.85&0.80\\ 
			BPOC&0.91&0.77&0.83&0.87&0.92&0.89&0.93&0.90&0.91\\ 
			INTENSIFIER&0.82&0.80&0.81&0.84&0.95&0.89&0.87&0.94&0.90\\ 
			DURATION&0.78&0.82&0.80&0.83&0.87&0.84&0.84&0.91&0.87\\ 
			NEGATION&0.83&0.90&0.86&0.84&0.89&0.86&0.83&0.93&0.88\\ 
			O&0.96&0.96&0.96&0.98&0.97&0.98&0.99&0.97&0.98\\ \hline
			MACRO&0.83&0.78&0.80&0.86&0.88&0.87&0.88&0.92&\textbf{0.90}\\ \hline
			
		\end{tabular}
	\end{center}
	\caption{Results of concept extraction from forum dataset using {\textbf{BERT+LSTM+CRF}} architecture. For the descriptions of {\textbf{BERT+LSTM+CRF}}, {\textbf{+DICT(1)}}, and {\textbf{+DICT(2)}} models see Subsection ~\ref{subsec:archmodels}. In all cases BERT parameters are frozen.} \label{tab:02}
\end{table*}    
\section{Methods}
\label{sec:methods}
A schematic of our architectures are shown in Figure ~\ref{fig:01}. Each architecture comprises Input Processing, Encoder Layer, and Decoder Layer. They differ in Input Processing unit, where we add a BERT layer, hence the architecture in Figure ~\ref{fig:01_a} is denoted as {\em LSTM+CRF} and the one in Figure ~\ref{fig:01_b} is denoted as {\em BERT+LSTM+CRF}. We now give details of these units as follows.

\subsubsection*{Input Processing}
\label{subsec:archinp}
The input processing unit consists of (i) input sequence, (ii) word piece tokenization, (iii) dictionary vector, (iv) embedding layer, and (v) frozen BERT layer. The units (i), (iii), and (iv) are common in both architectures where as (ii) and (v) are used with BERT+LSTM+CRF architecture. \\
\textit{\textbf{Input Sequence:}} An input sequence is either a sentence (from the forum dataset) or a complete tweet (from the Twitter dataset). The sequences are tokenized using  the GATE \cite{gate} software package.  For a given sequence, $S$ of length $l$, from this tokenization procedure we obtain $w_1,w_2,\ldots, w_l$ {\em tokens}. The vocabularies are constructed from the  unique tokens of these datasets.\\
\textit {\textbf{Dictionary Vector}}: The sequences of a post or a tweet is processed to construct a dictionary vector. The dictionary vector for token $w_i$ consists of 7 bits of information, where each bit is denoted as $d_i$, and represent either a dictionary or Unified Medical Language System (UMLS) semantic types. The example of a dictionary matrix for a selected sequence is shown in Figure ~\ref{fig:02}. Here, after tokenization, we processed the sequence using a NLP pipeline constructed using the GATE software. For dictionary/gazetteer matches we configure the pipeline for full match. We have five dictionaries in our pipeline. They are as follows:($d_1$) Symptom, ($d_2$) Severity, ($d_3$) Duration, ($d_4$) Intensifier, and ($d_5$) Negation. The dictionaries were built by analysing the posts while annotating them. We also utilized MetaMap, to map tokens to {\em Body Part, Organ, or Organ Component}, and {\em Sign or Symptom}, and {Disease or Syndrome} semantic concepts to represent $d_6$ and $d_7$ bits in our dictionary vector. Thus, for a given sequence, $S$, we collect $\vec{d_1},\vec{d_2},\ldots, \vec{d_l}$ vectors.\\
\textit{\textbf{BERT Tokenization}}: BERT transformer uses word piece tokenization procedure to tokenize a sentence. Here we denote the tokens of $S$ as $wp_1,wp_2,\ldots, wp_m$. Note that $l$ and $m$ may not match and $m>=l$. For example the word ``COVID`` is splited into two sub words [``CO``, ``\#VID``] in case of word piece tokenization. \\
\textit{\textbf{Embedding Layer}}: If a token is word pieced at position $i$, then the token and dictionary vector, $wi$ and $\vec{d_i}$, respectively, are repeated for the same number of times it is pieced. As a result, for the architecture in Figure ~\ref{fig:01_b}, the length of $S$ is extended to $m$. We collected pre-trained word embeddings $\vec{v_1}, \vec{v_2},\ldots, \vec{v_m}$ for the sequence $S$ from the GOOGLE news corpus \cite{mikolov2013}. If a word is not present in the vocabulary, the embedding is initialized randomly. Similarly the dictionary vectors are mapped to  $\vec{d_1},\vec{d_2},\ldots, \vec{d_m}$ for S.\\
\textit{\textbf{Frozen BERT Layer:}} For BERT+LSTM+CRF architecture in Figure ~\ref{fig:01_b}, we utilize BERT models by freezing their parameters and fed word pieced tokens into the layer before inputting them to the encoder layer. This produces contextual BERT vectors for a sequence and denoted as $\vec{b_1}, \vec{b_2}, \ldots, \vec{b_m}$. Specifically, we took representations from the last BERT layer.
\subsubsection*{Encoder Layer}
\label{subsec:archenc}
The encoder in our architecture is a two layered BiLSTM network which is 
similar to the architecture presented in \cite{lample-etal-2016-neural}. 
The {\em BiLSTM Layer 1} in the LSTM+CRF architecture is fed with pre-trained word vectors $\vec{v_1}, \vec{v_2},\ldots, \vec{v_l}$. For BERT+LSTM+CRF architecture, we feed concatenation of $\vec{v_1}, \vec{v_2},\ldots, \vec{v_m}$ and $\vec{b_1}, \vec{b_2}, \ldots, \vec{b_m}$. After processing, the first BiLSTM layer produces hidden representations $\vec{h_i^1}$. This representation is concatenated with $\vec{d_i}$ and fed into the second layer. Let the output of the second layer is $\vec{h_i^2}$. We also incorporated an attention layer ontop of the last hidden layer for LSTM+CRF architecture. We examined two types of attention: (i) self attention, and (ii) cross attention. In case of self attention, the query, key and value vectors come from the same hidden representations for each token. Whereas, in case of cross attention, the query is the last hidden representation of the LSTM which is deemed as the sentence representation, and the key and value vectors are each token`s hidden representations. We found that cross attention works well the LSTM+CRF architecture, however, the BERT+LSTM+CRF architecture are found to perform worse when attention is added. So we removed them from BERT+LSTM+CRF architecture.
\subsubsection*{Decoder Layer}
\label{subsec:archdec}
\label{subsec:02_2}
Our decoder layer is comprises a Softmax layer and a neural CRF. The hidden representations, $\vec{h_i^2}$, from the final encoder layer is fed into a Softmax layer to produce emission probabilities to a tag sequence. The emission probabilities are used with the neuro CRF to predict the final tag sequences. For implementation details, see \cite{lample-etal-2016-neural}.
\subsection{Models}
\label{subsec:archmodels}
From LSTM+CRF architecture we build the following models:
\begin{enumerate}
	\item LSTM+CRF: The {\em BiLSTM Layer 1} is initialised with pre-trained word vectors $\vec{v_i}$.
	\item +DICT(1): The {\em BiLSTM Layer 1} is initialised with the concatenation of dictionary and word vectors $\vec{d_i}$, $\vec{v_i}$, respectively.
	\item +DICT(2): The {\em BiLSTM Layer 1} is initialised with pre-trained word vectors $\vec{v_i}$ to produce $\vec{h_i^1}$. The dictionary vector $\vec{d_i}$ is concatenated with $\vec{h_i^1}$ and fed into {\em BiLSTM Layer 2}.
	
\end{enumerate}
From BERT+LSTM+CRF architecture we build the following models:
\begin{enumerate}
	\item BERT+LSTM+CRF: Concatenation of $\vec{v_i}$ and $\vec{b_i}$ are fed in to {\em BiLSTM Layer 1}.
	\item +DICT(1): Concatenation of $\vec{v_i}$, $\vec{b_i}$, and $\vec{d_i}$ are fed in to {\em BiLSTM Layer 1}.
	\item +DICT(2): Concatenation of $\vec{v_i}$ and $\vec{b_i}$ are fed in to {\em BiLSTM Layer 1} to produce $\vec{h_i^1}$. The dictionary vector $\vec{d_i}$ is concatenated with $\vec{h_i^1}$ and fed into {\em BiLSTM Layer 2}.
\end{enumerate}
\begin{table*}[h!] 
	\begin{center}
		
		\begin{tabular}[h!]{c|ccc|ccc|ccc}
			&\multicolumn{9}{c}{\textbf{Test datasets}} \\ \cline{2-10} 
			\textbf{Included Dictionary}&\multicolumn{3}{c|}{\textbf{Combined}}&\multicolumn{3}{c|}{\textbf{Our}}&\multicolumn{3}{c}{\textbf{Sarker}} \\ \cline{2-10} 
			{\bf Size} \%&P&R&$F_1$&P&R&$F_1$&P&R&$F_1$ \\ \hline
			0\%&1.00&0.83&0.90&1.00&1.00&1.00&0.63&0.72&0.67\\ 
			20\%&1.00&0.93&0.96&0.94&1.00&0.97&0.67&0.88&0.76\\ 
			40\%&1.00&0.96&0.98&0.92&1.00&0.96&0.68&0.94&0.79\\ 
			60\%&1.00&0.98&0.99&0.91&1.00&0.95&0.69&0.96&0.80\\ 
			80\%&1.00&1.00&1.00&0.89&1.00&0.94&0.70&1.00&0.82\\ 
			100\%&1.00&1.00&1.00&0.89&1.00&0.94&0.70&1.00&0.82\\ \hline
		\end{tabular}
		
	\end{center}
	\caption{Results of weakly supervised symptom extraction task using Our baseline {\textbf{LSTM+CRF+DICT(2)}} model  and for incremental additions from the Sarker dictionary.} \label{tab:03}
\end{table*}
\begin{table*}[h!]
	\begin{center}
		\begin{tabular}[h!]{c|ccc|ccc|ccc}
			&\multicolumn{9}{c}{\textbf{Test datasets}} \\ \cline{2-10} 
			\textbf{Included Dictionary}&\multicolumn{3}{c|}{\textbf{Combined}}&\multicolumn{3}{c|}{\textbf{Our}}&\multicolumn{3}{c}{\textbf{Sarker}} \\ \cline{2-10} 
			{\bf Size} \%&P&R&$F_1$&P&R&$F_1$&P&R&$F_1$ \\ \hline
			0\%&1.00&0.59&0.74&0.81&0.51&0.63&1.00&1.00&1.00\\ 
			20\%&1.00&0.88&0.94&0.88&0.86&0.87&0.77&1.00&0.87\\ 
			40\%&1.00&0.92&0.96&0.88&0.90&0.89&0.74&1.00&0.85\\ 
			60\%&1.00&0.92&0.96&0.88&0.89&0.89&0.73&0.97&0.84\\ 
			80\%&1.00&0.98&0.99&0.89&0.98&0.93&0.71&1.00&0.83\\ 
			100\%&1.00&1.00&1.00&0.89&1.00&0.94&0.70&1.00&0.82\\ \hline 
		\end{tabular}	
	\end{center}
	\caption{Results of weakly supervised symptom extraction task using Sarker baseline {\textbf{LSTM+CRF+DICT(2)}} model and for incremental additions from Our dictionary.}\label{tab:04} 
\end{table*}
\subsection{Transfer learning/Weak supervision}
For investigating how well our dictionaries based on the forum data can transfer to another dataset,  we focus on extracting symptoms only.
We introduce one  further dictionary that has been  developed by analysing the same Twitter dataset we are using in our forthcoming experiments which was  published by Sarker et al. \cite{sarker2020self}.  We  prune this dictionary by removing terms related to anxiety, stress \& general mental health symptoms and some phrases related to pyrexia or fever such as {\em 102 fever}, {\em 103+ fevers}, and {\em fever spiked to 107}. The number surrounding the term {\em fever} is annotated as {\em Severity} in our forum dataset.\\
To distinguish between the two dictionaries we call the forum built dictionary, {\em Our dictionary} and the Twitter based dictionary the {\em Sarker dictionary}.
\par From the Twitter dataset we extracted tweets which had at least one symptom. We did this by simply using Sarkar's dictionary to identify them. We removed 1000 tweets and annotate them which we used for our ground truth experiments. We used weak learning to train models in two ways:
\begin{enumerate}
	\item {\em Our base-line}: The train dataset is tagged using  Our dictionary.  
	\item {\em Sarker base-line}: The train dataset is tagged by the Sarker dictionary.  
\end{enumerate} 
We also looked at the effect of incrementally combining the dictionaries. Starting with  Our dictionary  we  include 20\% of Sarker's dictionary, which we  then use to tag the training data. We repeatedly add  a further 20\%  of Sarker's dictionary and tag the data again  until we have the union of both dictionaries. We repeat the process of tagging the training data starting with the Sarker dictionary  and incrementally including 20\% of Our dictionary.
For test set we not only evaluate on the ground truth but also on a  {\em weakly learnt test set} that is generated by tagging using the union of both dictionaries, i.e. Combined dictionaries. This latter test set is useful to see how well each individual dictionary can represent a dictionary that is generated from the combined datasets. For completeness we also look at the performance when we tag the test set using only the individual dictionaries separately.
\section{Experimental setup}
\label{sec:exp}
\subsection{Data}
\label{subsec:expdata}
We extracted 3000 posts related to COVID-19  from a patient social media forum  called Patient \cite{PI}, from this we randomly selected 500 social media posts to manually annotate. These posts were annotated with the class labels representing symptoms and the related concepts: (1) duration; (2) intensifier, which increases the level of symptom severity; (3)
severity; (4) negation, which denotes the presence or absence of the symptom or severity; and (5) affected body parts. The details of data collection procedure can be found in \cite{hasan2022monitoring}.
\par We collected tweets from the first 3 months of 2020 that contained at least one symptom, amounting to 36204 tweets, from a multilingual COVID-19 dataset published through the Github  \footnote{https://github.com/echen102/COVID-19-TweetIDs} repository by the authors of \cite{chen2020tracking}, and we manually annotated 1000 randomly selected tweets. 
\subsection{Settings}
\label{subsec:expset}
For both dataset we reported a 3-fold cross-validated $F_1$ scores. Training is done with the batch size of 16. The maximum sequence length for forum posts and tweets are 512 and 130, respectively. The number of hidden layers for each LSTM is 100. We use the Adam optimizer with a learning rate of 0.01 and a weight decay of 1e-5. The experiments were performed using the \texttt{transformers} library \cite{wolf2019huggingface} and all models were trained on an  NVIDIA Tesla P100. 
\begin{table*}[h!]
	\begin{center}
		\begin{tabular}[h!]{c|ccc|ccc|ccc}
			&\multicolumn{9}{c}{\textbf{Test datasets}} \\ \cline{2-10} 
			\textbf{Included Dictionary}&\multicolumn{3}{c|}{\textbf{Combined}}&\multicolumn{3}{c|}{\textbf{Our}}&\multicolumn{3}{c}{\textbf{Sarker}} \\ \cline{2-10} 
			{\bf Size} \%&P&R&$F_1$&P&R&$F_1$&P&R&$F_1$ \\ \hline
			0\%&1.00&0.82&0.90&1.00&1.00&1.00&0.63&0.72&0.67\\ 
			20\%&1.00&0.92&0.96&0.94&1.00&0.97&0.67&0.87&0.76\\  
			40\%&1.00&0.96&0.98&0.91&1.00&0.96&0.69&0.94&0.79\\  
			60\%&1.00&0.98&0.99&0.91&1.00&0.95&0.69&0.96&0.81\\  
			80\%&1.00&1.00&1.00&0.89&1.00&0.94&0.70&1.00&0.82\\  
			100\%&1.00&1.00&1.00&0.89&1.00&0.94&0.70&1.00&0.82\\ \hline  
		\end{tabular}
	\end{center}
	\caption{Results of weakly supervised symptom extraction task using Our baseline {\textbf{BERT+LSTM+CRF+DICT(2)}} model and for incremental additions of the Sarker dictionary. All experiments are performed using COVID-19 version of BERTweet.} \label{tab:05}
\end{table*}
\begin{table*}[h!]
	\begin{center}
		\begin{tabular}[h!]{c|ccc|ccc|ccc}
			&\multicolumn{9}{c}{\textbf{Test datasets}} \\ \cline{2-10} 
			\textbf{Included Dictionary}&\multicolumn{3}{c|}{\textbf{Combined}}&\multicolumn{3}{c|}{\textbf{Our}}&\multicolumn{3}{c}{\textbf{Sarker}} \\ \cline{2-10} 
			{\bf Size} \%&P&R&$F_1$&P&R&$F_1$&P&R&$F_1$ \\ \hline
			0\%&1.00&0.60&0.75&0.81&0.52&0.63&1.00&1.00&1.00\\ 
			20\%&1.00&0.88&0.94&0.87&0.86&0.86&0.77&1.00&0.87\\  
			40\%&1.00&0.92&0.96&0.88&0.90&0.89&0.75&1.00&0.85\\  
			60\%&1.00&0.93&0.96&0.88&0.92&0.90&0.74&1.00&0.85\\  
			80\%&1.00&0.98&0.99&0.89&0.98&0.93&0.71&1.00&0.83\\  
			100\%&1.00&1.00&1.00&0.89&1.00&0.94&0.70&1.00&0.82\\ \hline 
		\end{tabular}
	\end{center} 
	\caption{Results of weakly supervised symptom extraction task using Sarker baseline {\textbf{BERT+LSTM+CRF+DICT(2)}} model  and for incremental additions of Our dictionary. All experiments are performed using COVID-19 version of BERTweet.} \label{tab:06}
\end{table*}

\subsection{Results}
\label{subsec:ecpres}
\subsubsection*{Supervised Concept Extraction}
Results for the supervised experiments of the forum dataset are shown in Table ~\ref{tab:01} and ~\ref{tab:02} for LSTM+CRF and BERT+LSTM+CRF models, respectively. In both cases, incorporating the dictionary information into the input of the second BiLSTM layer, {\textbf{+DICT(2)}}, performs better than incorporating it into the first layer,{\textbf{+DICT(1)}}. Including BERT performs marginally better than not having it.   
\subsubsection*{Weak Supervision}
For weak supervision results we focus only on the {\textbf{+DICT(2)}} models since that was the best performing location for a dictionary in the supervised extraction experiments. Tables ~\ref{tab:03} and ~\ref{tab:04}  show the result from the LSTM+CRF model when Our and Sarker dictionary is used as the baseline, respectively. We note the final row from each table is the same, since the training labels are identical. We note also for the Combined test set, the labels for both the training and test are generated using the same dictionaries, hence  we get an $F_1$ of 1. 

Overall we see from Tables ~\ref{tab:03} and ~\ref{tab:04} that combining the dictionaries, even incrementally,  improves performance. Focussing on the first row and final column of Table ~\ref{tab:03} we see that our dictionary  performs favourably  ($F_1$ of 0.67) when the test data has been weakly labelled using a dictionary derived from that data  compared to when we swap the roles of the dictionaries, which can be seen in Table 4 first row  middle column ($F_1$ of 0.63).
\par
Tables ~\ref{tab:05} and ~\ref{tab:06} show results for when we include BERT features. We see that we achieve similar results which suggests that the information that BERT provides does not contribute much more than we already have through the static embedding and our dictionary features. 
\subsubsection*{Ground truth}
For the experiments involving the ground truth we found that replacing BERT with a COVID-19 version of BERTweet produced better results, which we report here in Tables ~\ref{tab:07} and ~\ref{tab:08}. As one would expect the results are lower than the experiments performed with weakly labelled test sets, nevertheless the performance is still good. Notably the finding that a pre-trained language model does not appreciably improve performance is observed here too. We present in Appendix A  several  tweets showing the ground truth and classifications using different dictionaries. 
\begin{table}[h!]
	\begin{center}	
		\begin{tabular}[h!]{l|llllll}  \hline
			&\multicolumn{3}{l}{\textbf{LSTM}}&\multicolumn{3}{l}{\textbf{BERTweet}} \\ \cline{2-7} 
			\textbf{\%}&P&R&$F_1$&P&R&$F_1$\\ \hline
			0\%&0.80&0.56&0.66&0.82&0.57&0.68\\ 
			20\%&0.82&0.65&0.72&0.83&0.65&0.73\\ 
			40\%&0.83&0.72&0.77&0.84&0.70&0.77\\ 
			60\%&0.82&0.74&0.78&0.84&0.73&0.78\\ 
			80\%&0.83&0.78&0.80&0.84&0.76&0.80\\ 
			100\%&0.83&0.78&0.81&0.84&0.76&0.80\\ \hline 
		\end{tabular}	
	\end{center}
	\caption{ Results of symptom extraction from the ground truth test set using Our base-line with incremental additions from Sarker dictionary. LSTM, and BERTweet correspond models with and without the language model, see main text.} \label{tab:07}
\end{table}
\begin{table}[h!]
	\begin{center}	
		\begin{tabular}[h!]{l|llllll}  \hline
			&\multicolumn{3}{l}{\textbf{LSTM}}&\multicolumn{3}{l}{\textbf{BERTweet}} \\ \cline{2-7} 
			\textbf{\%}&P&R&$F_1$&P&R&$F_1$\\ \hline
			0\%&0.92&0.60&0.72&0.92&0.57&0.70\\ 
			20\%&0.87&0.74&0.80&0.88&0.72&0.79\\ 
			40\%&0.86&0.75&0.80&0.87&0.73&0.80\\
			60\%&0.85&0.75&0.80&0.87&0.74&0.80\\
			80\%&0.84&0.78&0.81&0.85&0.76&0.80\\
			100\%&0.83&0.78&0.81&0.84&0.76&0.80\\ \hline 
		\end{tabular}
	\end{center}
	\caption{Results of symptom extraction from the ground truth test set using Sarker base-line with incremental additions from Our dictionary. LSTM, and BERTweet correspond models with and without the language model, see main text.} \label{tab:08}
\end{table}
\begin{table}[h]
	\centering
	\begin{tabular}[h!]{p{12cm}}  \hline
		Example 1\\ \hline 
		\\(a)\colorbox{green!20}{headache}, \colorbox{green!20}{fatigue}, \colorbox{green!20}{sore throat}, \colorbox{green!20}{cough} and \colorbox{green!20}{chest pressure} since sunday night. no \colorbox{green!20}{fever} though! but if its not covid, i dont know what it is. \\ \hline 
		\\(b) \colorbox{green!20}{headache}, \colorbox{green!20}{fatigue}, \colorbox{green!20}{sore throat}, \colorbox{green!20}{cough} and \colorbox{green!20}{chest pressure} since sunday night. no \colorbox{green!20}{fever} though! but if its not covid, i dont know what it is. \\ \hline 
		\\(c) \colorbox{green!20}{headache}, \colorbox{green!20}{fatigue}, \colorbox{green!20}{sore throat}, \colorbox{green!20}{cough} and \colorbox{green!20}{chest pressure} since sunday night. no \colorbox{green!20}{fever} though! but if its not covid, i dont know what it is. \\ \\ \hline 
	\end{tabular}
	\caption{Comparing models using examples. Here, (a), (b), and (c) denote ground truth, prediction from Sarker baseline, and prediction from Sarker baseline + Our dictionary, respectively} \label{tab:09}
	
\end{table}
\begin{table}[h]
	\centering
	\begin{tabular}[h!]{p{12cm}}  \hline
		Example 2 \\ \hline 
	\\
	(a) he left the house, got off sick bed, hardly able to stand, \colorbox{green!20}{muscle spasms}, not cognisant enough to remember speaking to pm, extremely low \colorbox{green!30}{blood oxygen levels} enough to be in hospital, broke lock down and self isolating rules and probable road traffic laws drove to hospital.\\ \\\hline 
	\\(b) he left the house, got off sick bed, hardly able to stand, \colorbox{green!20}{muscle spasms}, not cognisant enough to remember speaking to pm, extremely low \colorbox{red!20}{blood oxygen levels} enough to be in hospital, broke lock down and self isolating rules and probable road traffic laws drove to hospital.\\ \\ \hline 
	\\(c) he left the house, got off sick bed, hardly able to stand, \colorbox{green!20}{muscle spasms}, not cognisant enough to remember speaking to pm, extremely low \colorbox{green!20}{blood oxygen levels} enough to be in hospital, broke lock down and self isolating rules and probable road traffic laws drove to hospital.\\ \\ \hline 
	\end{tabular}
	\caption{Comparing models using examples. Here, (a), (b), and (c) denote ground truth, prediction from Sarker baseline, and prediction from Sarker baseline + Our dictionary, respectively} \label{tab:10}
\end{table}
\begin{table}[h]
	\centering
	\begin{tabular}[h!]{p{12cm}}  \hline
		Example 3 \\ \hline 
		\\
		(a) covid - 19 patients experience \colorbox{green!20}{loss of appetite}, \colorbox{green!20}{diarrhoea} and other \colorbox{green!20}{digestive symptoms}.\\ \\ \hline 
		\\(b) covid - 19 patients experience \colorbox{green!20}{loss of appetite}, \colorbox{green!20}{diarrhoea} and other \colorbox{red!20}{digestive symptoms.}\\ \\ \hline 
		\\(c) covid - 19 patients experience \colorbox{green!20}{loss of appetite}, \colorbox{green!20}{diarrhoea} and other \colorbox{green!20}{digestive symptoms}.\\ \\  
		\hline  
	\end{tabular}
	\caption{Comparing models using examples. Here, (a), (b), and (c) denote ground truth, prediction from Sarker baseline, and prediction from Sarker baseline + Our dictionary, respectively} \label{tab:11}
\end{table}
\begin{table}[h]
	\centering
	\begin{tabular}[h!]{p{12cm}}  \hline
	Example 4 \\ \hline
	\\ 
	(a) \colorbox{green!20}{smell} that? if not, you should probably call your doctor study finds \colorbox{green!20}{loss of taste and smell} can indicate covid-19.\\ \\ \hline 
	\\(b) \colorbox{red!20}{smell} that? if not, you should probably call your doctor study finds \colorbox{green!20}{loss of taste and smell} can indicate covid-19.\\ \\ \hline 
	\\(c) \colorbox{green!20}{smell} that? if not, you should probably call your doctor study finds \colorbox{green!20}{loss of taste and smell} can indicate covid-19.\\ \\
	\hline
	\end{tabular}
	\caption{Comparing models using examples. Here, (a), (b), and (c) denote ground truth, prediction from Sarker baseline, and prediction from Sarker baseline + Our dictionary, respectively} \label{tab:12}
\end{table}
\clearpage

\section{Discussion}
\label{sec:discussion}

We show some example tweets in Tables ~\ref{tab:09}, ~\ref{tab:10}, ~\ref{tab:11}, and ~\ref{tab:12}  with tagging results from our models. All the examples are taken from the LSTM model when Sarker dictionary is used as a base line and and for incremental additions from  our dictionary. The test data is labelled with the  ground truth. The Example 1 in Table ~\ref{tab:09} shows that the tweet contain COVID-19 symptoms such as {\em headache, fatigue, soar throat, and cough} which are common in both dictionaries. In the Example 2 of Table ~\ref{tab:10}, the concept {\em blood oxygen levels} does not have a presence in the Sarker dictionary. However, when it reaches 100\% with our dictionary, the model finds the symptom. Similarly, in Example 3 of Table ~\ref{tab:11}, with the addition of our dictionary the model finds out {\em digestive symptoms}. In Example 4 of Table ~\ref{tab:12}, we show that though {\em loss of taste and smell} exist in the Sarker dictionary, due to the longest match operation, the dictionary does tag the single word {\em smell} as symptom. However, since our dictionary has {\em smell} in it, the model is able to correctly find it when its coverage is increased.
\section{Conclusion}
Our experiments have shown that building small domain specific set of dictionaries can be beneficial for COVID-19 medical concept extraction. These dictionaries have the advantage that they are easy to produce and are interpretable. Moreover, models built using these dictionaries can generalize well and it is possible to transfer them to different datasets on a similar task. The results are encouraging in that a small domain specific set of dictionaries based on forum data can perform commensurately with BERTweet  on Twitter data when they are included as features in a model.


\section*{Acknowledgements}

{\em For the purposes of open access, the author has applied a CC BY public copyright licence to any author accepted manuscript version arising from this submission}



\bibliographystyle{unsrt}
 
\end{document}